\documentclass[10pt,twocolumn,letterpaper]{article}

\usepackage{wacv}              

\definecolor{wacvblue}{rgb}{0.21,0.49,0.74}
\usepackage[pagebackref,breaklinks,colorlinks,allcolors=wacvblue]{hyperref}
\usepackage{float}

\usepackage{setspace}
\usepackage[linesnumbered,ruled,vlined]{algorithm2e}
\usepackage{multirow}
\usepackage{amsthm}

\theoremstyle{remark}

\usepackage[table]{xcolor}
\definecolor{paleyellow}{HTML}{e1f3fc}
\definecolor{myblue}{HTML}{dfecf6}
\usepackage{graphicx}
\usepackage{amsmath}
\usepackage{amssymb}
\usepackage{booktabs}
\usepackage{multirow,makecell}
\usepackage{color}
\usepackage[table,dvipsnames]{xcolor}
\newcommand{\projectname}{ZO-SAM\xspace}
\usepackage{soul}

\title{ZO-SAM: Zero-Order Sharpness-Aware Minimization for Efficient Sparse Training}

\author{
\begin{tabular}{c}
Jie Ji$^{1}$ \quad
Gen Li$^{1}$ \quad
Kaiyuan Deng$^{2}$ \quad
Fatemeh Afghah$^{1}$ \quad
Xiaolong Ma$^{2}$ \\[0.5em]
$^{1}$Clemson University \quad
$^{2}$University of Arizona \\[0.3em]
{\tt\small jji@g.clemson.edu}
\end{tabular}
}

\begin{document}
\maketitle
\begin{abstract}
Deep learning models, despite their impressive achievements, suffer from high computational costs and memory requirements, limiting their usability in resource-constrained environments. Sparse neural networks significantly alleviate these constraints by dramatically reducing parameter count and computational overhead. However, existing sparse training methods often experience chaotic and noisy gradient signals, severely hindering convergence and generalization performance, particularly at high sparsity levels. 
To tackle this critical challenge, we propose \underline{Z}ero-\underline{O}rder \underline{S}harpness-\underline{A}ware \underline{M}inimization (\underline{ZO-SAM}), a novel optimization framework that strategically integrates zero-order optimization within the SAM approach. 
Unlike traditional SAM, ZO-SAM requires \textbf{only a single} backpropagation step during perturbation, selectively utilizing zero-order gradient estimations.
This innovative approach reduces the backpropagation computational cost by half compared to conventional SAM, significantly lowering gradient variance and effectively eliminating associated computational overhead. 
By harnessing SAM's capacity for identifying \textbf{flat minima}, ZO-SAM stabilizes the training process and accelerates convergence. 
These efficiency gains are particularly important in sparse training scenarios, where computational cost is the primary bottleneck that limits the practicality of SAM. 
Moreover, models trained with ZO-SAM exhibit improved robustness under distribution shift, further broadening its practicality in real-world deployments.
\end{abstract}

\section{Introduction}
\label{sec:intro}

Deep learning models have demonstrated impressive achievements across a wide variety of domains. Nevertheless, their substantial computational costs and extensive memory demands significantly hinder their deployment in resource-constrained environments, including edge devices, mobile applications, and scenarios involving large-scale pretraining. Sparse neural networks, which maintain only a small proportion of active weights, provide an attractive solution by drastically reducing parameter footprint and computation cost \cite{mocanu2018scalable, frankle2018lottery, lee2018snip, wang2019picking, tanaka2020pruning, evci2020rigging, yuan2021mest}. 

\begin{figure*}[t]
\centering
\includegraphics[width=1.0\textwidth]{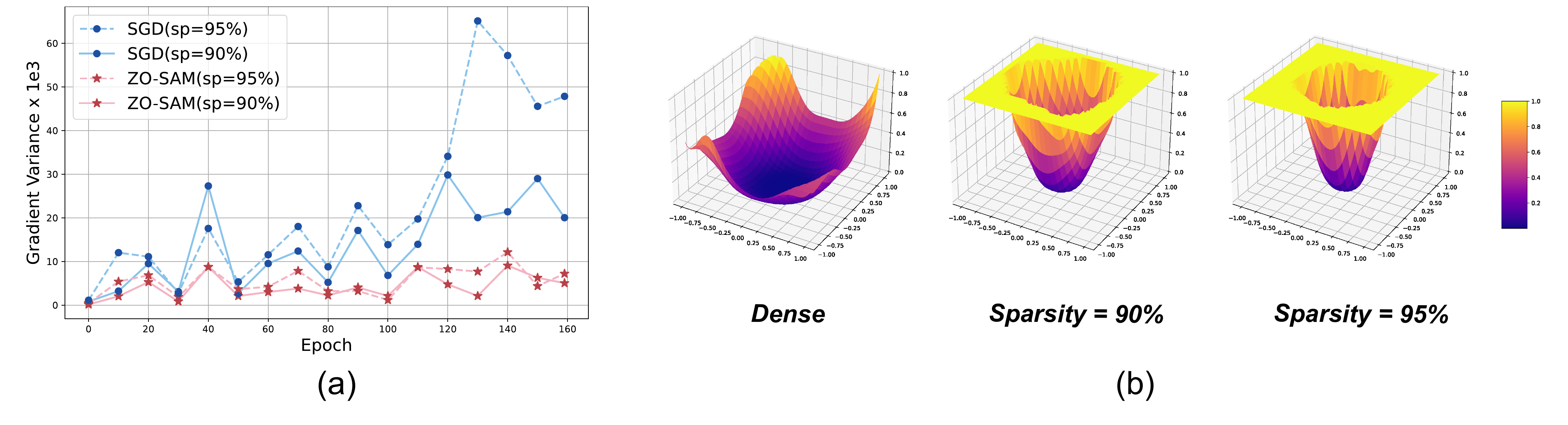}
\caption{(a) Comparison of Gradient Variance during sparse training at sparsity 90\% and 95\% between SGD and ZO-SAM; (b) Loss landscape of dense training, sparse training at sparsity 90\% and 95\%.}
\label{fig:loss_surface}
 \vspace{-0.2cm}
\end{figure*}

Despite their compelling advantages, current sparse training methodologies face critical challenges. Specifically, these approaches often rely on heuristic or specific metric strategies that compromise generalization, particularly at higher sparsity levels. Increasing sparsity exacerbates these limitations, making stable convergence and competitive accuracy progressively harder to achieve. In our investigation, we pinpoint \textbf{chaotic and noisy gradient signals} as the underlying cause of these learning difficulties. When a large fraction of weights is pruned, the remaining parameters bear a disproportionate burden, which disrupts the learning dynamics, limiting sparse models' convergence rate, generalization capability, and performance optima. This phenomenon is clearly illustrated in Figure \ref{fig:loss_surface} (a), which shows that traditional Stochastic Gradient Descent (SGD) exhibits highly erratic gradient fluctuations during sparse training, with increased sparsity further amplifying gradient variance. Such turbulent gradients cause the optimization trajectory to become inefficient and convoluted, impeding rapid convergence to loss minimum. Additionally, Figure \ref{fig:loss_surface} (b) demonstrates that introducing or increasing sparsity transforms the loss surface from a smooth, wide basin into a steeper, narrower landscape. This shift further confirms that sparsity induces noisier and more unstable gradient conditions, complicating effective gradient descent.

Sharpness-Aware Minimization (SAM) \cite{foret2020sharpness} has established itself as a powerful optimization technique to mitigate gradient instability by directing models toward \textit{flat minima} characterized by broader, smoother loss landscapes. Studies have shown that models trained with SAM achieve flatter, wider and  smoother loss landscapes, whereas models trained via standard methods often settle into steeper and narrower loss landscapes. Flat minima correspond to smooth loss surfaces, where small parameter changes—such as those induced by sampling noise—lead to only minor variations in the loss and its gradient. This property notably reduces gradient variance, stabilizing the training process and improving generalization. Mathematically speaking, if the gradient is Lipschitz continuous with a small Lipschitz constant, the fluctuations in gradient updates remain bounded, ensuring a more stable optimization trajectory. Given these benefits, integrating SAM into sparse training holds the promise of \textit{steering optimization processes towards solutions that are both more stable and more generalizable}. 

Nonetheless, SAM's significant computational overhead remains a critical limitation. 
Each optimization step requires \textit{two full backpropagations}, effectively \textbf{doubling the computational cost} compared to standard Stochastic Gradient Descent (SGD). 
While this extra cost might still be affordable in dense training, it becomes prohibitive in \textbf{sparse training scenarios}, where the motivation itself is to save computation and memory. 
In such resource-limited settings, the doubled overhead of SAM directly undermines its practicality, highlighting the urgent need for a more efficient alternative.

A key question arises: \textit{\ul{can SAM's structure be re-engineered to eliminate the necessity for \textbf{dual pass} of backpropagation per optimization step, thus avoiding the doubling of computational costs compared to SGD?}}

Zero-Order (ZO) Optimization presents a promising approach. Unlike traditional first-order methods, which depend on explicit gradient computation via backpropagation, ZO methods require \textit{zero extra} back propagation by estimate gradients using direct function evaluations. This characteristic makes ZO particularly advantageous in scenarios where explicit backpropagation is computationally costly or infeasible.
However, fully substituting first-order gradients with ZO estimates introduces additional approximation noise, potentially undermining update precision and overall training stability.

To overcome this limitation and effectively balance computational efficiency and optimization stability, we propose \textbf{Zero-Order SAM (ZO-SAM)}, a novel optimization framework that integrates Zero-Order (ZO) Optimization into SAM. Specifically, our ZO-SAM method strategically employs zero-order estimations solely during SAM's perturbation step, while maintaining precise first-order gradients for the subsequent parameter updates (illustrated in Figure \ref{fig:zosam}). This selective integration enables ZO-SAM to efficiently approximate perturbations with significantly reduced computational overhead while preserving the precision of first-order gradients essential for stable optimization. Consequently, ZO-SAM retains SAM's beneficial ability to guide models towards flatter minima, ensuring stable and generalizable optimization outcomes.

As evidenced in Figure \ref{fig:loss_surface} (a), ZO-SAM substantially decreases gradient variance compared to traditional SGD, particularly at high sparsity levels (90\% and 95\%), demonstrating enhanced optimization stability and improved convergence performance. Our comprehensive experiments illustrate that ZO-SAM surpasses existing sparse training techniques in accuracy and achieves comparable performance to SAM, all while significantly curbing computational demands. Consequently, our proposed method provides an optimal trade-off between computational cost and model performance, making it particularly suitable for resource-constrained deployment. 
Beyond efficiency and convergence, ZO-SAM also improves the \textbf{robustness} of sparse models. 
By reducing gradient variance and steering training toward flatter minima, ZO-SAM yields networks that generalize better and maintain higher accuracy under distribution shift, as further evidenced by our evaluations on loss surface visualizations, convergence rate analysis, comparisons with the SAM family, and robustness tests on CIFAR-10-C. In summary, the contributions of our work are as follow:


\begin{itemize} 
    \item We identify \textbf{chaotic and noisy gradient signals} as the critical obstacle impeding stable and accurate sparse neural network training. 
    \item We propose \textbf{ZO-SAM}, an innovative optimization framework tailored explicitly for sparse training scenarios, marking the first integration of \textbf{zero-order optimization} into the SAM framework. 
    \item Extensive empirical evaluations demonstrate that ZO-SAM achieves a superior \textbf{balance between computational efficiency, model accuracy, and robustness}, making it an effective and practical solution for sparse training, where controlling computational cost is particularly critical. 
\end{itemize}

\section{Related Works}
\label{sec:related_work}

\subsection{Sparse Training Methods}

Sparse neural network training methods aim to improve computational efficiency by enforcing many parameters to remain at zero throughout training, typically via binary masking of network weights \cite{evci2020rigging}.
\textbf{Static sparse training} determines the structure of the sparse network through the application of a pruning algorithm in the early stages of training. 
The lottery ticket hypothesis (LTH) \cite{frankle2018lottery,renda2020comparing,chen2020earlybert} uses iterative magnitude pruning (IMP), while methods such as SNIP~\cite{lee2018snip}, GraSP~\cite{wang2019picking}, and SynFlow~\cite{tanaka2020pruning} determine masks at initialization with gradient information. 
\textbf{Dynamic sparse training (DST)} instead starts with a random sparse network and adapts it throughout the training process to find a better sparse structure.
Sparse Evolutionary Training (SET)~\cite{mocanu2018scalable} prunes small magnitude weights and grows back randomly at the end of each training epoch. Deep R~\cite{bellec2018deep} uses a combination of stochastic parameter updates and dynamic sparse parameterization for training. Dynamic Sparse Reparameterization (DSR)~\cite{mostafa2019parameter} proposes to redistribute parameters between layers during training. 
Sparse Networks from Scratch (SNFS)~\cite{dettmers2019sparse} creates the sparse momentum algorithm, which finds the layers and weights that effectively reduce the error by using exponentially smoothed gradients (momentum).
In RigL~\cite{evci2020rigging}, the gradients of all the weights are computed when the model needs to be updated to grow new connections. Top-KAST~\cite{jayakumar2020top} proposes a scalable and performant sparse-to-sparse DST framework for maximum efficacy. MEST~\cite{yuan2021mest} designs a memory-economic sparse training framework targeting accurate and fast execution on edge devices. 
Static methods often suffer from poor convergence and limited generalization since many parameters remain inactive throughout training. 
DST alleviates these issues by updating masks, but usually requires long schedules or heavy hyperparameter tuning. 
Our proposed ZO-SAM complements these approaches by leveraging SAM to reduce gradient noise and encourage flatter minima, thereby improving convergence stability and generalization with controllable computational overhead.

\subsection{Zero-Order Optimization}
Zero-order optimization (ZO) methods optimize objective functions using only function evaluations without explicit gradient computations and enjoys provable convergence guarantees \citep{nesterov2017random,duchi2015optimal,liu2020primer}. ZO has gained significant attention for its success in various areas, including adversarial attack and defense \citep{chen2017zoo, tu2019autozoom,ye2018hessian,ilyas2018black,zhang2022robustify,verma2023certified, zhao2019design, hogan2018universal, shu2022zeroth}, visual prompting for transfer learning \citep{tsai2020transfer}, reinforcement learning \citep{vemula2019contrasting}, and LLM finetuning \citep{zhang2024revisiting, malladi2023fine, gautam2024variance}. Gradient estimation is critical in zeroth-order optimization methods, which aims to obtain the descent direction by sampling update directions and querying function evaluations. Extensive research has been conducted including smoothing and linear interpolation. The former methods smooth the objective function, causing a biased gradient estimation, while the latter often enjoys more accurate estimates, at the cost of large amounts of samples and queries at each iteration to update variables. Many methods have been proposed to reduce variance in ZO estimators \citep{chen2019zo,yuan2024new} and or faster convergence \citep{liu2019signsgd, an2025robust,qiu2024gradient,cai2022zeroth}.

\subsection{Sharpness-Aware Minimization}
Sharpness-aware minimization (SAM) improves generalization by explicitly optimizing towards flatter minima through a two-step perturbation-descent scheme \cite{foret2020sharpness}. However, its computational cost effectively doubles due to additional gradient calculations. Recent works have attempted to address this drawback. ESAM \cite{du2021efficient} reduces computational complexity by selectively perturbing only a subset of weights and training samples, significantly cutting costs while maintaining accuracy. LookSAM \cite{liu2022towards} exploits temporal consistency in SAM perturbation directions, periodically recomputing them to avoid redundant calculations. GSAM \cite{zhuang2022surrogate} enhances the generalization ability of SAM by adding an ascent step in the orthogonal direction to minimize the surrogate gap. Despite their advancements, these methods still require careful tuning of hyperparameters or scheduling mechanisms. In contrast, our proposed ZO-SAM inherently reduces computational overhead by eliminating one gradient backpropagation step per iteration through strategic integration of zero-order estimates, thus providing both simplicity and improved efficiency without sacrificing the flatness benefits of SAM.

\section{Method}
\label{sec:method}

\subsection{Sparse Training Formulation}

We consider a neural network parameterized by $\theta \in \mathbb{R}^d$ with a binary \emph{mask} $M \in \{0,1\}^d$ indicating which parameters are active (non-zero). Given a dataset $\mathcal{S}$ and loss function $\mathcal{L}_{\mathcal{S}}(\theta)$ (e.g., the average training loss), we define the \emph{sparse training problem} as:
\[
\begin{aligned}
\min_{\theta,\,M} \quad & \mathcal{L}_{\mathcal{S}}\bigl(f(x; \theta \odot M), y), \\
\text{subject to} \quad & \text{a sparsity constraint on } M.
\end{aligned}
\]
where ``$\odot$'' denotes the elementwise (Hadamard) product, \( f(x; \theta) \) is the neural network's prediction for input \( x \) with ground truth \( y \). Typically, we constrain the total number of active (non-zero) parameters to not exceed a budget $k$ (or equivalently enforce a target sparsity level $\alpha$). Formally,
\[
\|M\|_0 \;=\; \sum_{j=1}^d M_j \;\;\leq\; k,
\]
which ensures exactly (or at most) $k$ parameters are active. Sparse training methods may \emph{prune} (set to zero) some of the smallest-magnitude weights and optionally \emph{grow} new connections (activate new entries in $M$), thus evolving the mask over time while preserving overall sparsity.
 However, these methods often lead to high variance in gradient signals due to the dynamic nature of sparse weights, potentially resulting in unstable weight updates.

\subsection{Sharpness-Aware Minimization (SAM)}

Improving generalization in sparse models requires finding flat minima in the loss landscape, where the loss remains stable under small parameter perturbations. Sharpness-Aware Minimization (SAM)~\cite{foret2021sharpnessaware} addresses this by optimizing the worst-case loss in a neighborhood around the current parameters:
\begin{equation}
\min_{\theta} \max_{\|\epsilon\|_2 \leq \rho} \mathcal{L}(\theta + \epsilon)
\end{equation}
with \( \rho \) controlling the neighborhood size. SAM consists of two main steps:

\textbf{Perturbation Step:} Compute the adversarial perturbation:
\begin{equation}
\epsilon = \rho \frac{\nabla \mathcal{L}(\theta)}{\|\nabla \mathcal{L}(\theta)\|}.
\end{equation}

\textbf{Gradient Update Step:} Update the parameters using the gradient at the perturbed point:
\begin{equation}
\theta \leftarrow \theta - \eta \nabla \mathcal{L}(\theta + \epsilon),
\end{equation}
where \( \eta \) is the learning rate. While SAM offers desirable generalization properties, its computational demands restrict its applicability in sparse training scenarios, where efficiency is critical. 
\begin{figure}[t]
\centering
\includegraphics[width=0.9\columnwidth]{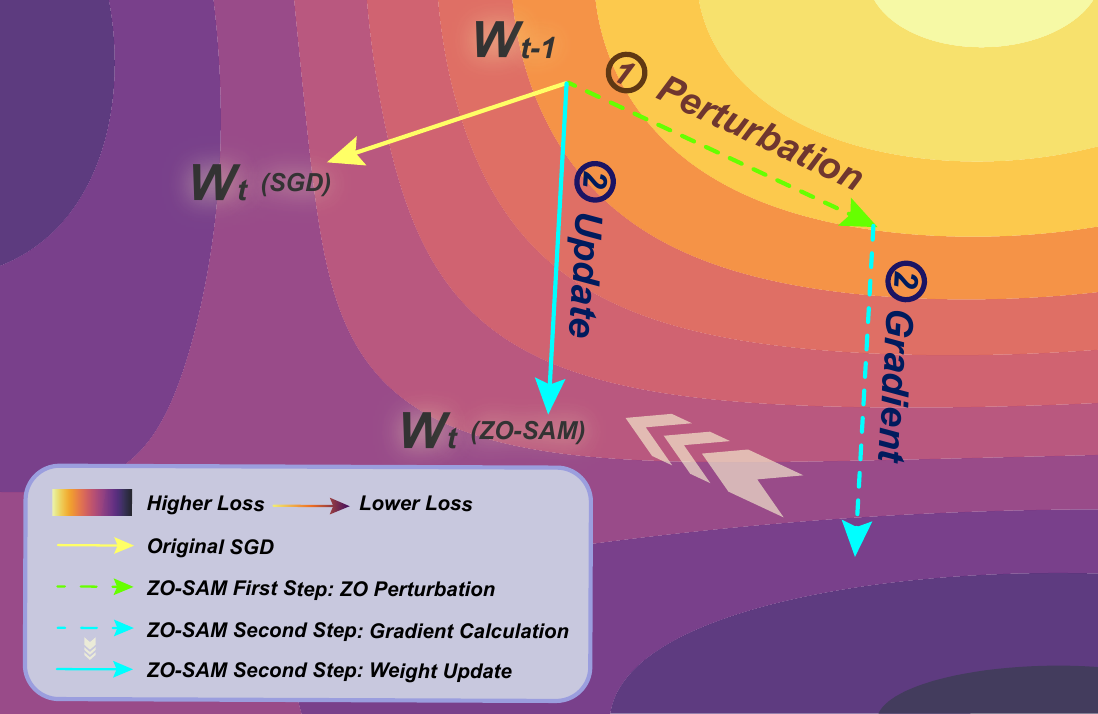}
\caption{Illustration of the optimization mechanism of ZO-SAM.}
\label{fig:zosam}
\end{figure}

\subsection{Zero-Order Optimization (ZO)}

Zero-Order (ZO) optimization methods estimate gradients using function evaluations rather than explicit backpropagation, offering an alternative approach for scenarios where gradient information is unavailable or computationally expensive. 

Two common ZO methods are:

\textbf{Random Gradient Estimation (RGE):} Approximates the gradient by averaging directional finite differences along \( m \) random directions:
\begin{equation}
\hat{\nabla} f(\theta) = \frac{1}{m} \sum_{i=1}^m \frac{f(\theta + \delta u_i) - f(\theta - \delta u_i)}{2\delta} u_i,
\end{equation}
with \( u_i \sim \mathcal{N}(0, I) \) and small step size \( \delta \). The use of random vectors \( u_i \) allows RGE to approximate the gradient with significantly fewer function evaluations than would be required by a coordinate-wise approach, particularly in high-dimensional parameter spaces.

Also, the aggregation of random directions helps avoid over-sensitivity to specific parameters, promoting broader exploration of the whole loss landscape and contributing to better generalization through smoother gradient updates.

\textbf{Coordinate-wise Gradient Estimation (CGE):} Approximates each gradient coordinate by evaluating perturbations along the corresponding axis:
\begin{equation}
\hat{\nabla} f(\theta)_j = \frac{f(\theta + \delta e_j) - f(\theta - \delta e_j)}{2\delta},
\end{equation}
where \( e_j \) is the unit vector in the \( j \)-th direction. While CGE may yield precise estimates, it requires \( d \) evaluations where \( d \) is the dimensionality of \( \theta \), , leading to extreme expensive computational cost for deep learning models with millions of parameters, making it less practical for large-scale sparse training scenarios.

In our proposed Zero-Order SAM (ZO-SAM) method, we adopt Random Gradient Estimation (RGE) instead of Coordinate-wise Gradient Estimation (CGE) due to both computational and theoretical considerations. Sparse training is often employed in environments with limited computational resources, where efficiency is critical. Since RGE requires only \( m \ll d \) function evaluations, it significantly reduces computational costs compared to CGE, which demands \( d \) evaluations. This efficiency is particularly valuable as it allows users to freely choose the value of \( m \)  based on their available resources and performance expectations. The flexibility of RGE aligns well with the design goals of ZO-SAM, which seeks to mitigate the computational overhead of Sharpness-Aware Minimization (SAM) while preserving its generalization benefits.

Beyond its computational advantages, RGE also provides theoretical benefits that make it particularly suitable for the perturbation step of SAM. This step is designed to explore the worst-case neighborhood of the loss landscape, where smooth gradient approximations play a crucial role. Unlike CGE, which computes gradients along individual coordinates and can introduce noise by overemphasizing specific sparse parameters, RGE’s random directional sampling offers a more holistic exploration of the landscape. This characteristic helps avoid sharp minima, leading to improved robustness and generalization, making RGE a natural fit for ZO-SAM.

A key design decision in ZO-SAM is the application of zero-order (ZO) estimation in the first step of SAM rather than in the second step. This choice is motivated by both computational and optimization considerations. In the perturbation step of SAM, minor inaccuracies in the gradient approximation are generally tolerable because the primary objective is to determine a suitable perturbation direction rather than to directly update model weights. By leveraging RGE in this step, ZO-SAM circumvents the need for a full backward pass, leading to substantial computational savings without compromising the benefits of sharpness-aware optimization. On the other hand, the gradient update step of SAM demands higher precision to ensure training stability. Retaining first-order gradients in this step allows ZO-SAM to maintain the balance between efficient exploration and precise exploitation, ensuring both optimization quality and stability.

Overall, the integration of RGE in ZO-SAM is a deliberate and principled design choice that enhances both computational efficiency and training stability. By applying zero-order estimation in the perturbation step while preserving first-order gradients for model updates, ZO-SAM achieves an optimal trade-off between computational feasibility and performance effectiveness. This hybrid approach ensures that ZO-SAM remains a robust and practical solution for Sharpness-Aware Optimization in sparse training scenarios, making it particularly well-suited for resource-constrained environments.
\begin{table*}[t]
\small
\centering
\caption{Test accuracy (\%) of pruned ResNet-32 on CIFAR-10/100.}
\label{tab:cifar}
\resizebox{1.0\textwidth}{!}{
\begin{tabular}{l ccc ccc}
\cmidrule[\heavyrulewidth](lr){1-7}
 \multirow{1}{*}{Datasets}     & \multicolumn{3}{c}{\multirow{1}{*}{CIFAR-10}} & \multicolumn{3}{c}{\multirow{1}{*}{CIFAR-100}}  \\ 
 \cmidrule(lr){1-7}
Pruning ratio   & 90\%      & 95\%     & 98\%     
     & 90\%      & 95\%     & 98\%       \\ 
\cmidrule(lr){1-1}
\cmidrule(lr){2-4}
\cmidrule(lr){5-7}

\textbf{ResNet-32}  
& 94.58 (Dense) &  & 
& 74.89 (Dense) &  & 
\\
\cmidrule(lr){1-1}
\cmidrule(lr){2-4}
\cmidrule(lr){5-7}
LT~[\citenum{frankle2018lottery}] 
& 92.31 & 91.06 & 88.78 
& 68.99 & 65.02 & 57.37 
\\

LT+ \projectname (ours) 
& \textbf{92.69$\pm$0.05} (0.38$\uparrow$) & \textbf{91.67$\pm$0.10} (0.61$\uparrow$) & \textbf{89.46$\pm$0.15} (0.68$\uparrow$)
& \textbf{69.44$\pm$0.06} (0.45$\uparrow$) & \textbf{65.59$\pm$0.13} (0.57$\uparrow$) & \textbf{57.87$\pm$0.14} (0.50$\uparrow$)
\\ \cmidrule(lr){1-7}
SNIP~[\citenum{lee2018snip}]
& 92.59$\pm$0.10 & 91.01$\pm$0.21 & 87.51$\pm$0.31
& 68.89$\pm$0.45 & 65.02$\pm$0.69 & 57.37$\pm$1.43
\\

SNIP+ \projectname (ours) 
& \textbf{93.38$\pm$0.12} (0.79$\uparrow$) & \textbf{91.69$\pm$0.17} (0.68$\uparrow$) & \textbf{88.22$\pm$0.21} (0.71$\uparrow$)
& \textbf{69.54$\pm$0.23} (0.65$\uparrow$) & \textbf{65.90$\pm$0.49} (0.88$\uparrow$) & \textbf{58.29$\pm$0.47} (0.92$\uparrow$)
\\ \cmidrule(lr){1-7}
GraSP~[\citenum{wang2019picking}]
& 92.38$\pm$0.21  & 91.39$\pm$0.25  & 88.81$\pm$0.14
& 69.24$\pm$0.24  & 66.50$\pm$0.11  &  58.43$\pm$0.43
\\

GraSP+ \projectname (ours) 
& \textbf{92.97$\pm$0.24} (0.59$\uparrow$) & \textbf{91.93$\pm$0.11} (0.54$\uparrow$) & \textbf{89.68$\pm$0.13} (0.87$\uparrow$)
& \textbf{70.11$\pm$0.19} (0.87$\uparrow$) & \textbf{67.31$\pm$0.28} (0.81$\uparrow$) & \textbf{59.60$\pm$0.20} (1.17$\uparrow$)
\\
\midrule

SET~[\citenum{mocanu2018scalable}] 
 & 92.30 & 90.76 & 88.29
 & 69.66 & 67.41 & 62.25
\\

SET+ \projectname (ours) 
& \textbf{93.14$\pm$0.17} (0.84$\uparrow$) & \textbf{91.55$\pm$0.18} (0.79$\uparrow$) & \textbf{88.98$\pm$0.23} (0.69$\uparrow$)
& \textbf{70.43$\pm$0.23} (0.77$\uparrow$) & \textbf{68.40$\pm$0.15} (0.99$\uparrow$) & \textbf{63.77$\pm$0.23} (1.52$\uparrow$)
\\ \cmidrule(lr){1-7}
DSR~[\citenum{mostafa2019parameter}] 
 & 92.97 & 91.61 & 88.46
 & 69.63 & 68.20 & 61.24
\\

DSR+ \projectname (ours) 
& \textbf{93.74$\pm$0.19} (0.77$\uparrow$) & \textbf{92.29$\pm$0.22} (0.68$\uparrow$) & \textbf{89.21$\pm$0.14} (0.75$\uparrow$)
& \textbf{70.40$\pm$0.15} (0.77$\uparrow$) & \textbf{69.07$\pm$0.15} (0.87$\uparrow$) & \textbf{62.23$\pm$0.17} (0.99$\uparrow$)
\\ \cmidrule(lr){1-7}

RigL~[\citenum{evci2020rigging}] 
& 93.07 & 91.83 & 89.00
& 70.34 & 68.22 & 64.07
\\

RigL+ \projectname (ours) 
& \textbf{93.66$\pm$0.12} (0.59$\uparrow$) & \textbf{92.21$\pm$0.19} (0.38$\uparrow$) & \textbf{90.61$\pm$0.13} (1.61$\uparrow$)
& \textbf{72.88$\pm$0.16} (2.54$\uparrow$) & \textbf{70.58$\pm$0.13} (2.36$\uparrow$) & \textbf{65.17$\pm$0.09} (1.10$\uparrow$)

\\ \cmidrule(lr){1-7}
MEST~[\citenum{yuan2021mest}] 
 & 92.56$\pm$0.07 & 91.15$\pm$0.29 & 89.22$\pm$0.11
 & 70.44$\pm$0.26 & 68.43$\pm$0.32 & 64.59$\pm$0.27
\\

MEST + \projectname (ours) 
& \textbf{93.50$\pm$0.11} (0.94$\uparrow$) & \textbf{91.88$\pm$0.09} (0.73$\uparrow$) & \textbf{91.53$\pm$0.17} (2.31$\uparrow$)
& \textbf{72.20$\pm$0.14} (1.76$\uparrow$) & \textbf{70.30$\pm$0.09} (1.87$\uparrow$) & \textbf{66.01$\pm$0.29} (1.42$\uparrow$)
\\ 

\cmidrule[\heavyrulewidth](lr){1-7}
\\
 \vspace{-0.4cm}
 \end{tabular}}
  \label{tab:cifar10_resnet}
\end{table*}

\subsubsection{Algorithm}

\begin{figure}[H]
\centering
\includegraphics[width=1.0\columnwidth]{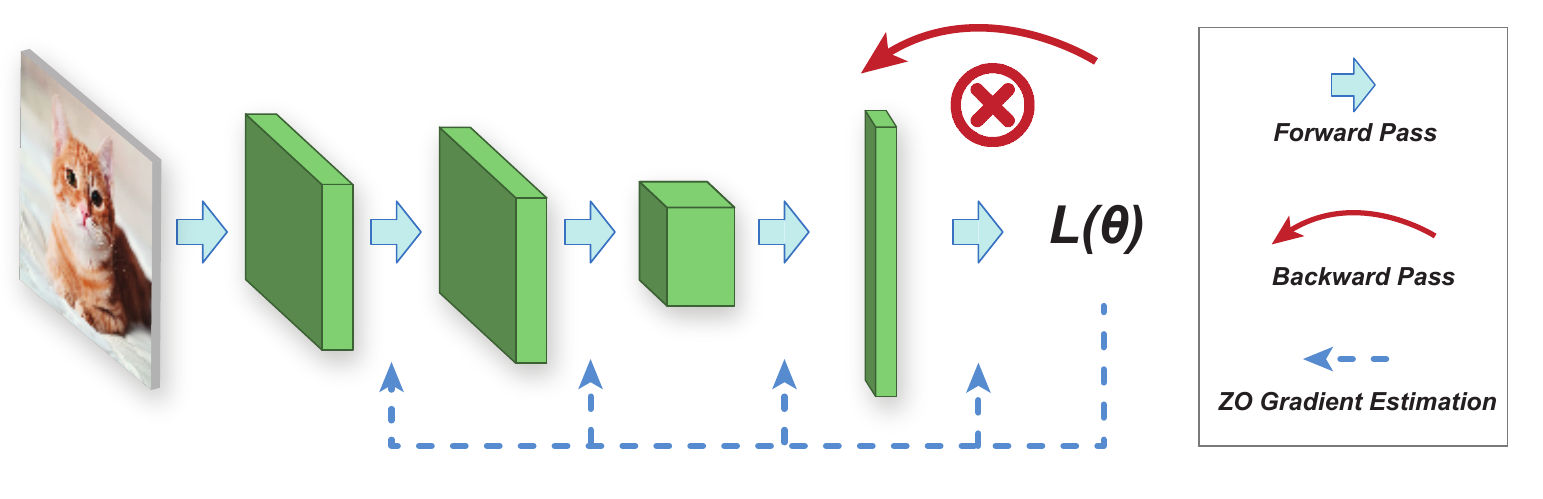}
\caption{Overview of perturbation step in ZO-SAM.}
\label{fig:overview}
\end{figure}

\textbf{Perturbation Step with Zero-Order Gradient:} Instead of computing the perturbation direction using first-order gradients, ZO-SAM employs RGE to approximate the gradient:
\begin{equation}
\epsilon = \rho \frac{\hat{\nabla} \mathcal{L}(\theta)}{\|\hat{\nabla} \mathcal{L}(\theta)\|}
\end{equation}
where
\begin{equation}
\hat{\nabla} \mathcal{L}(\theta) = \frac{1}{m} \sum_{i=1}^{m} \frac{\mathcal{L}(\theta + \delta u_i) - \mathcal{L}(\theta - \delta u_i)}{2\delta} u_i.
\end{equation}
As shown in Figure \ref{fig:overview}, in the perturbation step of ZO-SAM, backward pass is no longer needed, achieving a significant reduction in computational costs. From Figure \ref{fig:zosam}, we can see that this is  represented by the light green arrow labeling it is the perturbation step.

\textbf{Gradient Update Step with First-Order Precision:} The update step uses the precise first-order gradient to maintain stability and convergence:
\begin{equation}
\theta \leftarrow \theta - \eta \nabla \mathcal{L}(\theta^*(\epsilon)),
\end{equation}
where \( \theta^*(\epsilon) = \theta + \epsilon \). From Figure \ref{fig:zosam}, we can see that this is shown by the light blue arrow labeling gradient calculation and weight update. This choice ensures that the model benefits from SAM's ability to find flat minima, critical for generalization in sparse models.

The ZO-SAM algorithm for Sparse Training is outlined in Algorithm~\ref{alg:sparse-zo-sam} in appendix.

\begin{figure}[H]
\centering
\includegraphics[width=0.9\columnwidth]{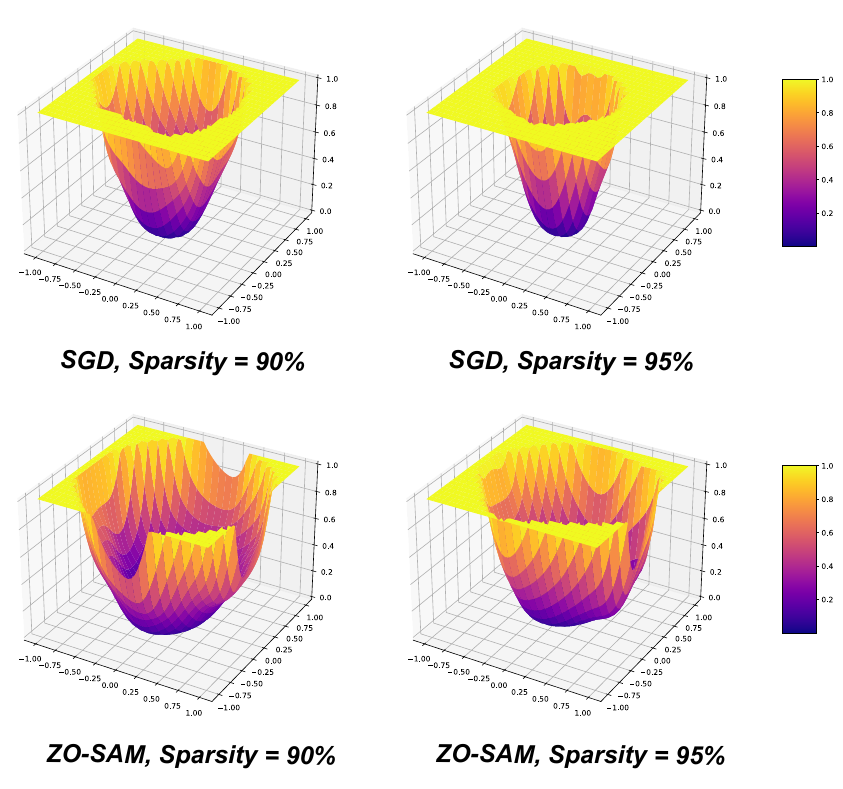}
\caption{Loss surface improvement at various sparsity levels.}
\label{fig:loss_surface_compare}
\end{figure}
\section{Experiments}
\label{sec:experiments}


We conduct comprehensive experiments to evaluate \projectname across \textbf{architectures} (ResNet-32~\cite{wang2019picking}, ResNet-50~\cite{he2016deep}, WRN-28-10~\cite{zagoruyko2016wide}, DeiT-Tiny/Small~\cite{touvron2021training}), \textbf{datasets} (CIFAR-10/100~\cite{krizhevsky2009learning}, CIFAR-10-C~\cite{hendrycks2019benchmarking}, ImageNet-1K~\cite{russakovsky2015imagenet}), and \textbf{sparse regimes} (90\%, 95\%, 98\%).
We cover \textbf{static} methods (LTH~\cite{frankle2018lottery}, SNIP~\cite{lee2018snip}, GraSP~\cite{wang2019picking}) and \textbf{dynamic} methods (SET~\cite{mocanu2018scalable}, DSR~\cite{mostafa2019parameter}, RigL~\cite{evci2020rigging}, MEST~\cite{yuan2021mest}).
For the SAM family, we compare \textbf{SAM}~\cite{foret2021sharpnessaware}, \textbf{ESAM}~\cite{du2021efficient}, \textbf{LookSAM (LS)}~\cite{liu2022towards}, and \textbf{GSAM}~\cite{zhuang2022surrogate} alongside \textbf{ZO-SAM}.


All of our experiments are performed on NVIDIA 4$\times$ A6000 GPUs. We repeat training experiments for 3 times and report the mean and standard deviation of the accuracy. 

\subsection{Accuracy Improvement for State-of-the-art Sparse Training Methods}

\textbf{ResNet on CIFAR-10/100.}
To validate the universal applicability of \projectname, we assess its performance across a variety of established sparse training methods at varying sparsity levels. The results on ResNet-32 is shown in Table \ref{tab:cifar10_resnet}, and the results on ResNet-50 is shown in Table \ref{tab:cifar10_resnet50} in appendix. For each baseline method, we perform training with \projectname at 90\%, 95\% and 98\% sparsity, and compare the accuracy with the original accuracy.  
Notably, our method consistently improves the performance across all pruning ratios, with significant gains observed in both datasets. For instance, in the CIFAR-10 dataset, our method boosts the accuracy of the previous sparse training method in ResNet-32, ranging from 0.38\% to 2.31\%. Similarly, on CIFAR-100, our method also delivers substantial improvements, with accuracy gains ranging from 0.45\% to 2.54\% in ResNet-32 depending on the baseline method and pruning ratio. Table~\ref{tab:cifar10_resnet50} in the appendix demonstrates that our method, \projectname, consistently enhances the accuracy of various pruning techniques across different pruning ratios on ResNet-50. Notably, we achieve accuracy improvements of up to \textbf{1.04}\% on CIFAR-10 and \textbf{1.56}\% on CIFAR-100. These results highlight the effectiveness of our approach in enhancing the performance of sparse training frameworks, even at high sparsity levels, and demonstrate its potential as a valuable tool for improving pruning strategies in deep learning. 

\textbf{Transformer on ImageNet-1K.} 
We further evaluate \projectname on transformer-based architectures using DeiT~\cite{touvron2021training} as the backbone. Following the experiment setup in SViTE~\cite{chen2021chasing},  a representative sparse training method for transformers, we train DeiT-Tiny and DeiT-Small on ImageNet-1K and compare different dynamic sparse training strategies with and without our method. Across both backbones, shown in Table~\ref{tab:transformer}, ZO-SAM consistently enhances performance: on DeiT-Tiny with 50\% sparsity, accuracy improves by up to \textbf{1.14}\%, while on DeiT-Small with 70\% sparsity, the gain reaches \textbf{1.17}\%.


\begin{table}
\caption{Test accuracy (\%) of pruned DeiT on ImageNet-1K}
    \centering
    \begin{tabular}{l|c c}
    \toprule
      Model   & DeiT-Small &DeiT-Tiny\\
      \cmidrule(lr){1-3}
      Dense & 79.90 & 72.20  \\
      \cmidrule(lr){1-3}
      Pruning ratio & 70\% & 50\%\\
      \cmidrule(lr){1-3}
      RigL~[\citenum{evci2020rigging}]  & 77.99 &70.02\\
      RigL+ZO-SAM & \textbf{79.16} (1.17$\uparrow$)&\textbf{70.79} (0.77$\uparrow$)\\
      \cmidrule(lr){1-3}
      MEST~[\citenum{yuan2021mest}]  & 77.15 &69.69\\
      MEST+ZO-SAM & \textbf{78.04} (0.89$\uparrow$)& \textbf{70.41} (0.72$\uparrow$)\\
      \cmidrule(lr){1-3}
      SViTE~[\citenum{chen2021chasing}]&78.18 &70.18\\
      SNIP+ZO-SAM & \textbf{79.31} (1.13$\uparrow$) &\textbf{71.32} (1.14$\uparrow$)\\
      \bottomrule
    \end{tabular}
    \label{tab:transformer}
\end{table}

\subsection{Loss Surface Visualization}

We visualize and compare the loss surfaces resulting from sparse training with and without \projectname, as illustrated in Figure~\ref{fig:loss_surface_compare}. Specifically, we conduct experiments utilizing the sparse training method MEST~\cite{yuan2021mest} at sparsity ratios of 90\% and 95\% on the CIFAR-10 dataset with ResNet-32. Observing these results, it becomes clear that higher sparsity ratios lead to steeper and narrower loss basins, indicating more chaotic and noisy gradients and weight updates. In contrast, when incorporating \projectname under identical sparsity conditions, the loss surface displays notably wider and smoother basins. This suggests that \projectname effectively stabilizes gradient calculations and weight updates, promoting a smoother loss trajectory throughout the sparse network training process and achieving improving generalization ability.

\subsection{Convergence Rate Comparison}

\begin{figure}[t]
\centering
\includegraphics[width=0.9\columnwidth]{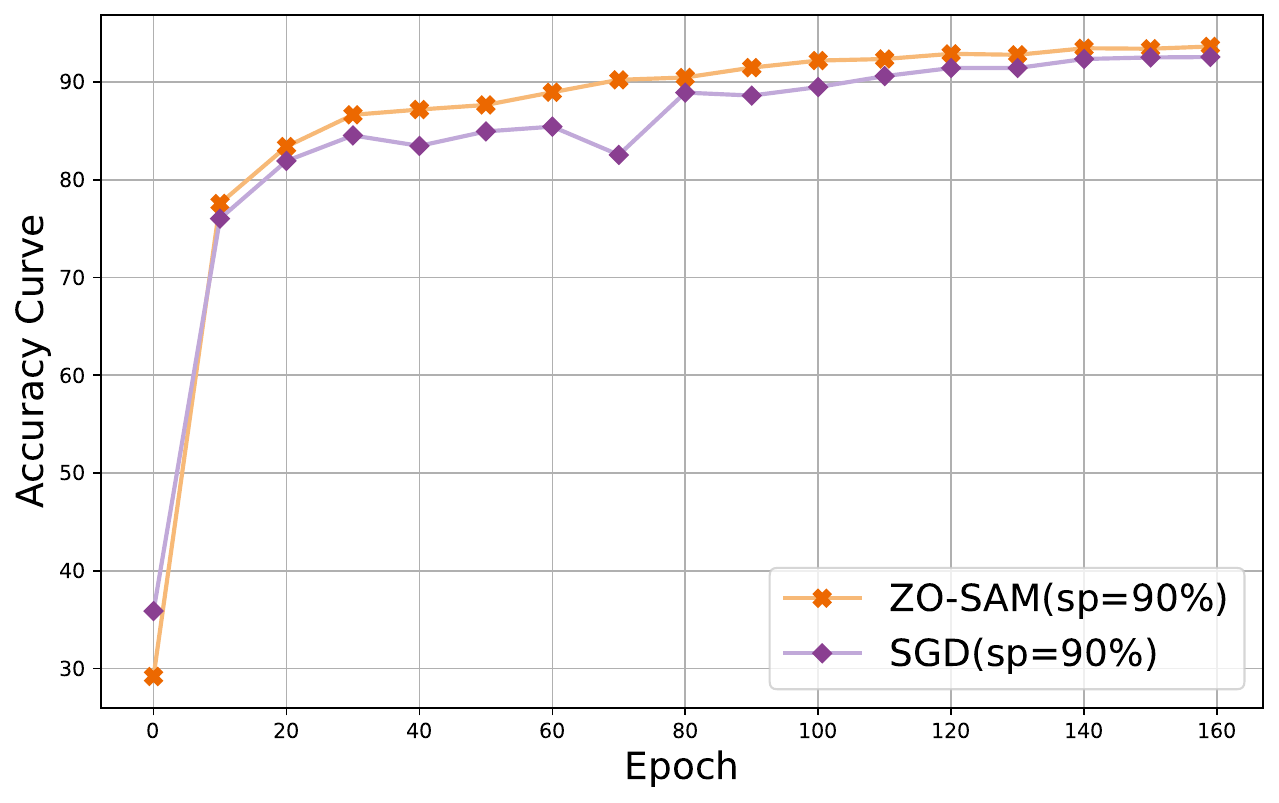}
\caption{Convergence Speed Comparison at sparsity level 90\%.}
\label{fig:acc_curve90}
\end{figure}

\begin{figure}[t]
\centering
\includegraphics[width=0.9\columnwidth]{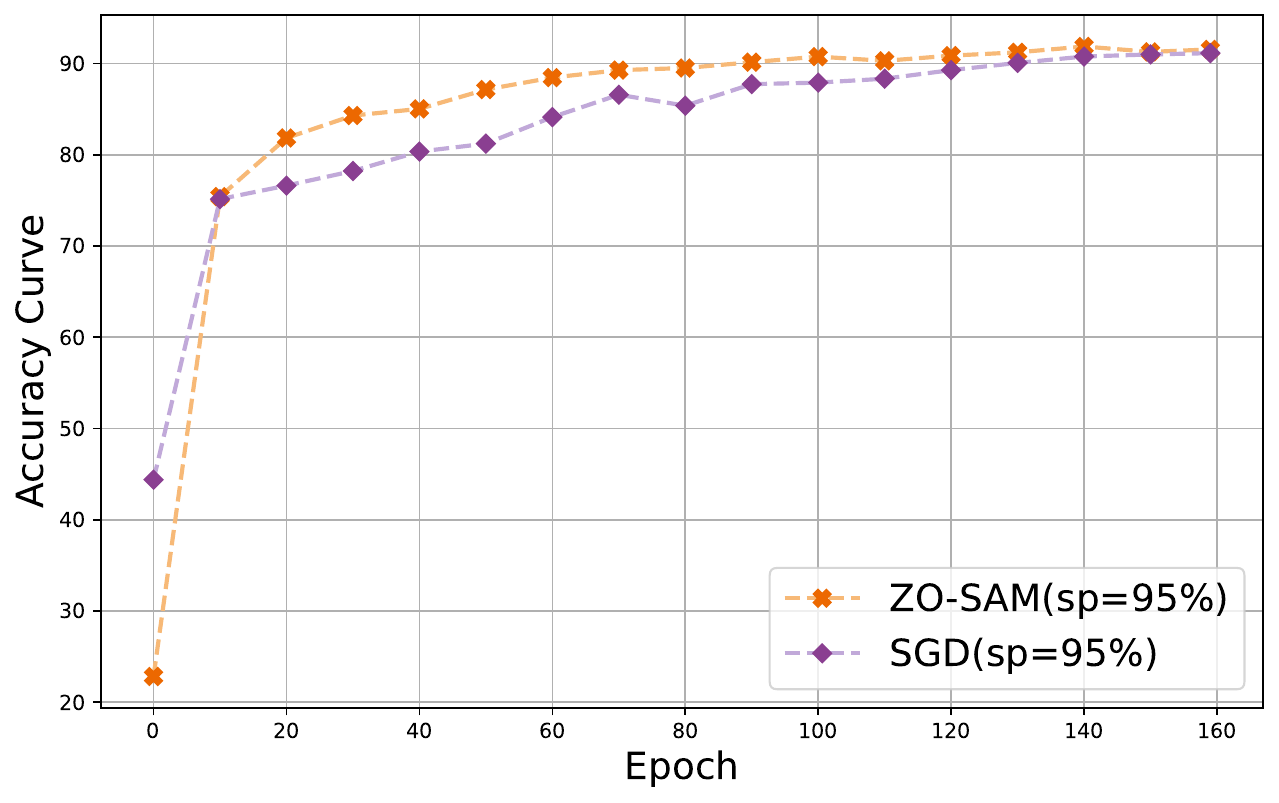}
\caption{Convergence Speed Comparison at sparsity level 95\%.}
\label{fig:acc_curve95}
\end{figure}



As shown in Figures~\ref{fig:acc_curve90} and \ref{fig:acc_curve95}, \projectname converges substantially faster than standard first-order baselines such as SGD. 
For example, while SGD typically requires more than 130 epochs to approach its final accuracy, \projectname reaches comparable performance within only about 120 epochs. 
Importantly, this improvement is achieved with just a \textbf{single backward pass per iteration}, highlighting the efficiency advantage of our design. 

To make convergence speed comparisons more explicit, we report in Table~\ref{tab:converge} the epoch at which each method first attains $90\%$ accuracy, where a lower epoch count indicates faster convergence.



\begin{table*}[t]
\linespread{1.0}
\small
\centering
\caption{Epochs to reach 90\% accuracy on CIFAR-10 (lower is better).}
\scalebox{1.0}{
\begin{tabular}{c|ccccc}
\toprule 
CIFAR-10, MEST, \textbf{Acc}=$90\%$ & SGD & ESAM & LookSAM(k=5) &GSAM & ZO-SAM\\
\midrule 
sp=0.9, Epoch\# & 104 &  75 & 79  &84& \cellcolor{myblue}\textbf{70}\\
sp=0.95, Epoch\# & 131 & 92 & 94  &113& \cellcolor{myblue}\textbf{88}\\
\bottomrule
\end{tabular}}
\label{tab:converge}
\end{table*}

\subsection{Feature Map Comparison}

\begin{table*}[t]
\centering
\caption{Test accuracy (\%) and computation efficiency (image/sec) of ResNet-32 and WideResNet-28-10 under 90\% sparsity levels on CIFAR-10 and CIFAR-100. Comparison includes SGD, original SAM, efficient SAM variants, and our proposed methods.}
\resizebox{0.98\textwidth}{!}{
\begin{tabular}{c|c|c|c|c|c|c|c}
\toprule
\multirow{2}{*}{\textbf{Method}} & \multirow{2}{*}{\textbf{Sparsity}} & \multicolumn{3}{c|}{\textbf{ResNet-32}} & \multicolumn{3}{c}{\textbf{WRN-28-10}} \\
\cmidrule(lr){3-8}
& & CIFAR-10 & CIFAR-100 & Image/sec & CIFAR-10 & CIFAR-100 & Image/sec \\
\midrule
SGD & - & 94.58 & 74.89 & 5673.95 & 96.41 & 81.63 & 752.30 \\
\midrule
SAM & 90\% & 93.77 & 72.64 & 2704.84 (47.67\%) & 95.37 & 80.14 & 354.94 (47.18\%)\\
ESAM~\cite{du2021efficient} & 90\% & 93.67 & 72.49 & 2297.23 (40.48\%) & 95.31 & 80.23 & 305.76 (40.64\%) \\
LS(k=5)~\cite{liu2022towards} & 90\% & 93.42 & 72.01 & 3980.62 (70.14\%) & 94.89 & 79.30 & 542.63 (72.13\%) \\
LS(k=10)~\cite{liu2022towards} & 90\% & 93.03 & 71.62 & 4272.22 (75.29\%) & 94.17 & 78.84 & 593.02 (78.83\%) \\
GSAM~\cite{zhuang2022surrogate}&90\% &93.72&72.90& 2701 (47.60\%)
&95.40& 80.28&340.19(45.22\%)\\
\rowcolor{myblue}
ZO-SAM (Ours) & 90\% & 93.50 & 72.20 & \textbf{4349.53 (76.67\%)} & 95.02 & 79.91 & \textbf{576.01 (76.57\%)} \\
\bottomrule

\end{tabular}}
\label{tab:different_sam_all}
\end{table*}

\begin{figure}[h]
\centering
\includegraphics[width=1.0\columnwidth]{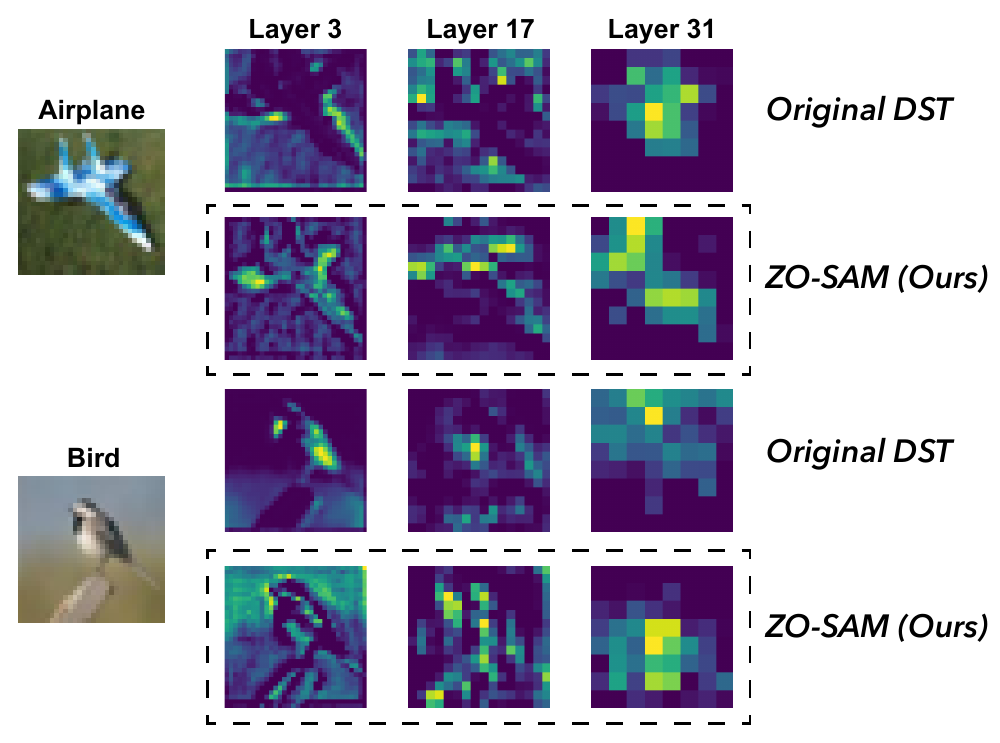}
\caption{Comparison of feature maps for different layers between Original DST and ZO-SAM under sparsity 90\%. }
\label{fig:featuremap}
\end{figure}

We compare feature maps at layers 3, 17, and 31 of ResNet-32 on representative samples (airplane, bird). In Figure \ref{fig:featuremap}, we show side-by-side visualizations for \textbf{ZO-SAM (Ours)} versus \textbf{Original DST}. We select these specific layers to capture early, mid-level, and deeper activations: layer 3 typically reveals edge-like responses, layer 17 exposes more intermediate texture patterns, and layer 31 shows higher-level semantic cues.

For both airplane and bird images, our method produces \textbf{crisper and more localized activations} than the baseline. In the early layer (layer 3), edges are more coherently outlined, suggesting the model is focusing cleanly on object boundaries. By the mid-layer (layer 17), texture representations appear less noisy, indicating improved stability. At the deep layer (layer 31), features are well-grouped around the main object regions, implying stronger semantic alignment. Conversely, the baseline feature maps show scattered or hazy patterns, indicating higher variance in internal representations.

These qualitative observations support our quantitative findings: by removing the first backward pass and relying on zero-order approximations, ZO-SAM retains sharpness-aware advantages—leading to flatter minima and improved generalization—while still learning clear, decisive feature representations across the network’s depth. Hence, the feature map comparison offers tangible evidence that our approach better captures the critical parts of each image and maintains robust activations throughout the layers.

\subsection{Comparison with SAM family}
In table~\ref{tab:different_sam_all}, we compare the test accuracy and computational efficiency (image per second) of ResNet-32 and WideResNet-28-10 under 90\% sparsity using MEST sparse training on CIFAR-10 and CIFAR-100. Among these methods, ZO-SAM offers a strong balance, achieving 93.50\% and 72.20\% accuracy on ResNet-32 while improving efficiency to 4349.53 images per second, significantly outperforming SAM. On WRN-28-10, ZO-SAM maintains high accuracy (95.02\% and 79.91\%) while achieving a processing speed of 576.01 images per second. 
While SAM and GSAM yield slightly higher accuracy in some cases, their throughput is much lower (around 60\% of \projectname). 
LookSAM (k=10) achieves higher efficiency on WRN-28-10, but its accuracy lags behind. 
Overall, these results highlight that \projectname consistently \textbf{balances accuracy and efficiency}, making it a practical choice for sparse training in resource-constrained settings.

\subsection{Robustness Improvement by ZO-SAM}

Since sparse models are widely deployed in real-world scenarios where test data may not match the training distribution, it is important to evaluate their robustness under distribution shift. 
We therefore assess robustness on CIFAR-10-C~\cite{hendrycks2019benchmarking}, which contains the same test images as CIFAR-10 but applies 19 common corruptions at 5 severity levels. 
Our experiments train ResNet-32 with SNIP at 90\% sparsity and compare SAM-family variants with \projectname. 
For each method, we report accuracy on the clean CIFAR-10 test set and on the corrupted CIFAR-10-C test set, and compute $\Delta = \text{Clean Acc} - \text{CIFAR-10-C Acc}$, where a smaller value indicates stronger robustness. 
As shown in Table~\ref{tab:robust}, \projectname improves corrupted-set accuracy by \textbf{3.10\%} over SNIP and reduces $\Delta$ accordingly, demonstrating that our approach yields more robust sparse models under distribution shift.

\begin{table}[h!]
\small
\centering
\captionof{table}{CIFAR-10 vs. CIFAR-10-C under 90\% sparsity with SNIP.}
\label{tab:robust}
\scalebox{0.9}{
\begin{tabular}{l|ccc}
\toprule 
Methods & CIFAR-10 (\%) & CIFAR-10-C (\%) & $\Delta\downarrow$ \\ 
\midrule
SNIP~\cite{lee2018snip} & 92.59 & 59.60 & 32.99\\ 
\midrule
SNIP + ESAM~\cite{du2021efficient} & 93.40 (0.81$\uparrow$) & 61.19 (1.59$\uparrow$) & 32.21\\
SNIP + LS(k=5)~\cite{liu2022towards} & 93.31 (0.72$\uparrow$) & 60.41 (0.81$\uparrow$) & 32.90 \\
SNIP + LS(k=10)~\cite{liu2022towards} & 92.97 (0.38$\uparrow$) & 60.08 (0.48$\uparrow$) & 32.89 \\
SNIP + GSAM~\cite{zhuang2022surrogate} &93.71 (1.12$\uparrow$) & 61.65 (2.05$\uparrow$) & 32.1\\
\rowcolor{myblue}
SNIP + \textbf{\projectname} & 93.38 (0.79$\uparrow$) & \textbf{62.70} (3.10$\uparrow$) & \textbf{30.68} \\
\bottomrule
\end{tabular}}
\end{table}

\section{Conclusion}
In this paper, we introduce Zero-Optimization Sharpness-Aware Minimization (\textbf{ZO-SAM}), an innovative method that transforms the computationally demanding SAM into an efficient framework by eliminating the requirement for dual backward passes.
ZO-SAM effectively addresses the trade-off between computational efficiency and optimization stability, offering a practical solution for integrating SAM into sparse training scenarios where resources are limited. Our proposed method significantly reduces computational overhead while preserving the desirable properties of flat minima, thereby enhancing training stability, generalization performance, and robustness under distribution shift.


{
    \small
    \bibliographystyle{ieeenat_fullname}
    \bibliography{main}
}

\newpage
\appendix
\counterwithin{figure}{section}
\counterwithin{table}{section}
\clearpage
{\huge \bf Appendix}

\begin{table*}[!t]
\small
\centering
\vspace{-0.5em}
\captionof{table}{Test accuracy (\%) of pruned ResNet-50 on CIFAR-10/100.}
\vspace{-0.5em}
\resizebox{1.0\textwidth}{!}{
\begin{tabular}{l ccc ccc}
\cmidrule[\heavyrulewidth](lr){1-7}
 \multirow{1}{*}{Datasets}     & \multicolumn{3}{c}{\multirow{1}{*}{CIFAR-10}} & \multicolumn{3}{c}{\multirow{1}{*}{CIFAR-100}}  \\ 
 \cmidrule(lr){1-7}
Pruning ratio   & 90\%      & 95\%     & 98\%     
     & 90\%      & 95\%     & 98\%       \\ 
\cmidrule(lr){1-1}
\cmidrule(lr){2-4}
\cmidrule(lr){5-7}

\textbf{ResNet-50}  
& 95.23 (Dense) &  & 
& 78.42 (Dense) &  & 
\\
\cmidrule(lr){1-1}
\cmidrule(lr){2-4}
\cmidrule(lr){5-7}

SNIP~[\citenum{lee2018snip}]
& 92.65$\pm$0.12 & 90.86$\pm$0.17 & 87.02$\pm$0.35
& 73.14$\pm$0.32 & 69.25$\pm$0.37 & 62.49$\pm$0.32
\\
SNIP+ \projectname (ours) 
& \textbf{92.98$\pm$0.09} (0.33$\uparrow$) & \textbf{91.47$\pm$0.12} (0.61$\uparrow$) & \textbf{87.92$\pm$0.14} (0.90$\uparrow$)
& \textbf{73.64$\pm$0.23} (0.50$\uparrow$) & \textbf{69.97$\pm$0.49} (0.72$\uparrow$) & \textbf{63.79$\pm$0.47} (1.30$\uparrow$)
\\ \cmidrule(lr){1-7}
GraSP~[\citenum{wang2019picking}]
& 92.47$\pm$0.17  & 91.32$\pm$0.15  & 88.77$\pm$0.24
& 73.28$\pm$0.21  & 70.29$\pm$0.13  &  62.12$\pm$0.33
\\
GraSP+ \projectname (ours) 
& \textbf{92.89$\pm$0.17} (0.42$\uparrow$) & \textbf{91.85$\pm$0.12} (0.53$\uparrow$) & \textbf{89.47$\pm$0.11} (0.70$\uparrow$)
& \textbf{73.79$\pm$0.16} (0.51$\uparrow$) & \textbf{70.98$\pm$0.24} (0.69$\uparrow$) & \textbf{63.27$\pm$0.20} (1.15$\uparrow$)
\\
\midrule

SET~[\citenum{mocanu2018scalable}] 
 & 94.47 & 94.12 & 87.14
 & 76.22 & 75.81 & 73.25
\\
SET+ \projectname (ours) 
& \textbf{94.81$\pm$0.13} (0.34$\uparrow$) & \textbf{94.55$\pm$0.21} (0.43$\uparrow$) & \textbf{88.18$\pm$0.29} (1.04$\uparrow$)
& \textbf{76.82$\pm$0.27} (0.60$\uparrow$) & \textbf{76.20$\pm$0.25} (0.39$\uparrow$) & \textbf{74.27$\pm$0.31} (1.02$\uparrow$)
\\ \cmidrule(lr){1-7}
DSR~[\citenum{mostafa2019parameter}] 
 & 94.55 & 94.17 & 87.46
 & 76.63 & 74.93 & 73.06
\\
DSR+ \projectname (ours) 
& \textbf{94.71$\pm$0.20} (0.16$\uparrow$) & \textbf{94.42$\pm$0.21} (0.25$\uparrow$) & \textbf{87.96$\pm$0.14} (0.50$\uparrow$)
& \textbf{76.94$\pm$0.13} (0.31$\uparrow$) & \textbf{75.67$\pm$0.15} (0.74$\uparrow$) & \textbf{74.01$\pm$0.17} (0.95$\uparrow$)
\\ \cmidrule(lr){1-7}

RigL~[\citenum{evci2020rigging}] 
& 94.43 & 94.21 & 93.14
& 77.34 & 75.22 & 74.87
\\
RigL+ \projectname (ours) 
& \textbf{94.61$\pm$0.13} (0.18$\uparrow$) & \textbf{94.41$\pm$0.19} (0.20$\uparrow$) & \textbf{93.51$\pm$0.19} (0.37$\uparrow$)
& \textbf{77.79$\pm$0.16} (0.45$\uparrow$) & \textbf{76.78$\pm$0.13} (1.56$\uparrow$) & \textbf{75.93$\pm$0.09} (1.06$\uparrow$)

\\ \cmidrule(lr){1-7}
MEST~[\citenum{yuan2021mest}] 
 & 94.06$\pm$0.09 & 93.95$\pm$0.18 & 92.87$\pm$0.12
 & 77.01$\pm$0.21 & 74.93$\pm$0.38 & 74.00$\pm$0.25
\\
MEST + \projectname (ours) 
& \textbf{94.37$\pm$0.10} (0.31$\uparrow$) & \textbf{94.08$\pm$0.09} (0.13$\uparrow$) & \textbf{93.53$\pm$0.17} (0.66$\uparrow$)
& \textbf{77.95$\pm$0.13} (0.94$\uparrow$) & \textbf{76.02$\pm$0.09} (1.09$\uparrow$) & \textbf{75.21$\pm$0.29} (1.21$\uparrow$)
\\ 

\cmidrule[\heavyrulewidth](lr){1-7}
\\
 \end{tabular}}
  \label{tab:cifar10_resnet50}
\end{table*}
\section{Algorithm}

\begin{algorithm}[htb]
\caption{Sparse Zero-Order SAM (Sparse ZO-SAM)}
\label{alg:sparse-zo-sam}
\KwIn{Initial parameter $\theta_0$, initial mask $M_0$, total iterations $T$, step sizes $\{\eta_t\}_{t=0}^{T-1}$, perturbation radii $\{\rho_t\}_{t=0}^{T-1}$, finite-difference step $\delta$, number of random directions $m$, weight decay parameter $\lambda$, dataset $\mathcal{S}$.} 
\For{$t \leftarrow 0$ \KwTo $T-1$}{
  \textbf{Apply the current mask:}
  \[
    \theta_t \leftarrow \theta_t \odot M_t \quad (\text{elementwise product}).
  \]
  \textbf{Zero-order gradient estimate:}
    \begin{align*}
  &\text{Sample } m \text{ random vectors } \{u_i\}_{i=1}^m \sim \mathcal{N}(0,I). \\
  &\hat{g}_t \leftarrow \frac{1}{m} \sum_{i=1}^m \frac{1}{2\delta} \Bigl[ L_{\mathcal{S}}\bigl((\theta_t + \delta u_i) \odot M_t\bigr) \\
  &\quad - L_{\mathcal{S}}\bigl((\theta_t - \delta u_i) \odot M_t\bigr) \Bigr] u_i.
\end{align*}
  
  \textbf{Compute the perturbation:}
  \[
    \epsilon_t \leftarrow \rho_t \,\frac{\hat{g}_t}{\|\hat{g}_t\|}.
  \]
  \textbf{Set the perturbed parameter:}
  \[
    \tilde{\theta}_t \leftarrow (\theta_t + \epsilon_t) \odot M_t.
  \]
  \textbf{Compute the first-order gradient at the perturbed parameter:}
  \[
    g_t \leftarrow \nabla L_{\mathcal{S}}\bigl(\tilde{\theta}_t\bigr).
  \]
  \textbf{Update parameters with weight decay:}
  \[
    \theta_{t+1} \leftarrow \bigl(\theta_t - \eta_t\,\bigl(g_t + \lambda\,\theta_t\bigr)\bigr) \odot M_t.
  \]
}
\Return $\theta_T$.
\end{algorithm}

\section{Results of ResNet-50 on CIFAR-10/100}

Table~\ref{tab:cifar10_resnet50} reports the detailed results of applying \projectname to ResNet-50 on CIFAR-10 and CIFAR-100 across 90\%, 95\%, and 98\% sparsity. We place this table in the appendix for completeness, as the main text focuses on ResNet-32 due to space constraints. 
Consistent with the observations in the main paper, \projectname consistently improves the accuracy of multiple sparse training baselines on ResNet-50. 
These results further support the conclusion that our method generalizes well across different architectures and sparsity levels.

\

\end{document}